\documentclass[letterpaper, 10 pt, conference]{ieeeconf}  

\IEEEoverridecommandlockouts               

\def\BibTeX{{\rm B\kern-.05em{\sc i\kern-.025em b}\kern-.08em
T\kern-.1667em\lower.7ex\hbox{E}\kern-.125emX}}

\usepackage{amsmath}
\usepackage{amsfonts}
\usepackage{amssymb}
\usepackage{amsbsy}
\usepackage{bbm}
\usepackage{algorithm}

\usepackage{xcolor}

\usepackage{xspace}

\usepackage{graphicx}
\usepackage{subfigure}
\usepackage{wrapfig}

\usepackage{booktabs} 
\usepackage{subfiles}
\usepackage{multicol}
\usepackage{latexsym} 
\usepackage{array}

\usepackage[bookmarks=true]{hyperref}
\let\labelindent\relax
\usepackage{enumitem}
\usepackage{subcaption} 
\usepackage[font=small,labelfont=bf]{caption}

\usepackage{times}

\usepackage{multicol}
\usepackage{amsmath,amssymb,amsfonts}
\usepackage{booktabs}
\usepackage{graphicx}
\usepackage{xcolor}
\usepackage{bbm}
\usepackage{nicefrac}
\usepackage{comment}
\usepackage{algorithm}
\usepackage{algpseudocode}

\newcommand{\sref}[1]{Sec. \ref{#1}}
\newcommand{\figref}[1]{Fig. \ref{#1}}
\newcommand{\tabref}[1]{Table \ref{#1}}

\definecolor{wine}{RGB}{204, 0, 102}
\definecolor{ocean}{RGB}{13, 121, 202}
\definecolor{light_ocean}{RGB}{18, 178, 235}
\definecolor{dark_ocean}{RGB}{10, 89, 148}
\definecolor{grey}{RGB}{170, 170, 170}
\definecolor{light-grey}{RGB}{220, 220, 220}
\definecolor{dark_gray}{rgb}{0.2, 0.2, 0.2} 
\definecolor{med-grey}{rgb}{0.3, 0.3, 0.3} 
\definecolor{grape}{RGB}{112,48,160}
\definecolor{aqua}{RGB}{52,172,139}
\definecolor{dark_aqua}{RGB}{35,115,93}
\definecolor{dark_orange}{RGB}{216,92,0}
\definecolor{vibrant_orange}{RGB}{255, 102, 0}
\definecolor{vibrant_blue}{RGB}{14, 120, 255}
\definecolor{vibrant_pink}{RGB}{255, 0, 104}
\definecolor{dark_red}{RGB}{122, 0, 0}
\definecolor{dark_green}{RGB}{0, 92, 34}
\definecolor{sunset}{RGB}{233, 108, 20}

\newcommand{\para}[1]{\medskip\noindent\textbf{#1. }}

\newcommand{\nosafety}{\textcolor{black}{\textbf{NoSafety}}\xspace}
\newcommand{\robust}{\textcolor{gray}{\textbf{Robust-RA}}\xspace}
\newcommand{\ssa}{\textcolor{grape}{\textbf{SSA}}\xspace}
\newcommand{\marginal}{\textcolor{aqua}{\textbf{Marginal-RA}}\xspace}
\newcommand{\ours}{\textcolor{orange}{\textbf{SLIDE}}\xspace}

\newcommand{\human}{\mathcal{H}}
\newcommand{\robot}{\mathcal{R}}

\newcommand{\state}{x}

\newcommand{\stateSpace}{\mathcal{S}}
\newcommand{\stateSpaceR}{\mathcal{S}_\robot}
\newcommand{\stateSpaceH}{\mathcal{S}_\human}
\newcommand{\qSpace}{\mathcal{Q}}

\newcommand{\ctrl}{u}
\newcommand{\aR}{\ctrl_\robot}
\newcommand{\aH}{\ctrl_\human}
\newcommand{\ctrlSett}{U}
\newcommand{\ctrlSet}{\mathcal{\ctrlSett}}
\newcommand{\ctrlSetH}{\ctrlSet_\human}
\newcommand{\ctrlSetR}{\ctrlSet_\robot}
\newcommand{\aRtraj}{\mathbf{\ctrl}_\robot}
\newcommand{\aHtraj}{\mathbf{\ctrl}_\human}

\DeclareMathOperator*{\argmin}{argmin}

\newcommand{\safeSet}{\mathcal{S}^\shield}

\usepackage{fontawesome5}
\newcommand{\policy}{\pi}
\newcommand{\shield}{*}

\newcommand{\fallback}{\policy^{\shield}_\robot}

\title{\LARGE \bf 
Robots that Learn to Safely Influence via \\ Prediction-Informed Reach-Avoid Dynamic Games
}

\author{Ravi Pandya, \quad Changliu Liu, \quad Andrea Bajcsy
\thanks{Authors are with the Robotics Institute at Carnegie Mellon University, Pittsburgh, Pennsylvania, \tt\small \{rapandya, cliu6, abajcsy\}@andrew.cmu.edu}
}

\begin{document}

\maketitle
\thispagestyle{plain}
\pagestyle{plain}

\begin{abstract}
Robots can influence people to accomplish their tasks more efficiently: autonomous cars can inch forward at an intersection to pass through, and  tabletop manipulators can go for an object on the table first. However, a robot's ability to influence can also compromise the safety of nearby people if naively executed. 
In this work, we pose and solve a novel robust reach-avoid dynamic game which enables robots to be maximally influential, but only when a safety backup control exists. 
On the human side, we model the human's behavior as goal-driven but conditioned on the robot's plan, enabling us to capture influence. 
On the robot side, we solve the dynamic game in the joint physical and belief space, enabling the robot to reason about how its uncertainty in human behavior will evolve over time. 
We instantiate our method, called SLIDE (Safely Leveraging Influence in Dynamic Environments), in a high-dimensional (39-D) simulated human-robot collaborative manipulation task solved via offline game-theoretic reinforcement learning. 
We compare our approach to a robust baseline that treats the human as a worst-case adversary, a safety controller that does not explicitly reason about influence, and an energy-function-based safety shield. 
We find that SLIDE consistently enables the robot to leverage the influence it has on the human when it is safe to do so, ultimately allowing the robot to be less conservative while still ensuring a high safety rate during task execution.
Project website: \url{https://cmu-intentlab.github.io/safe-influence/}
\end{abstract}

\IEEEpeerreviewmaketitle

\section{Introduction}

\begin{figure}[t!]
    \centering
    \includegraphics[width=0.9\columnwidth]{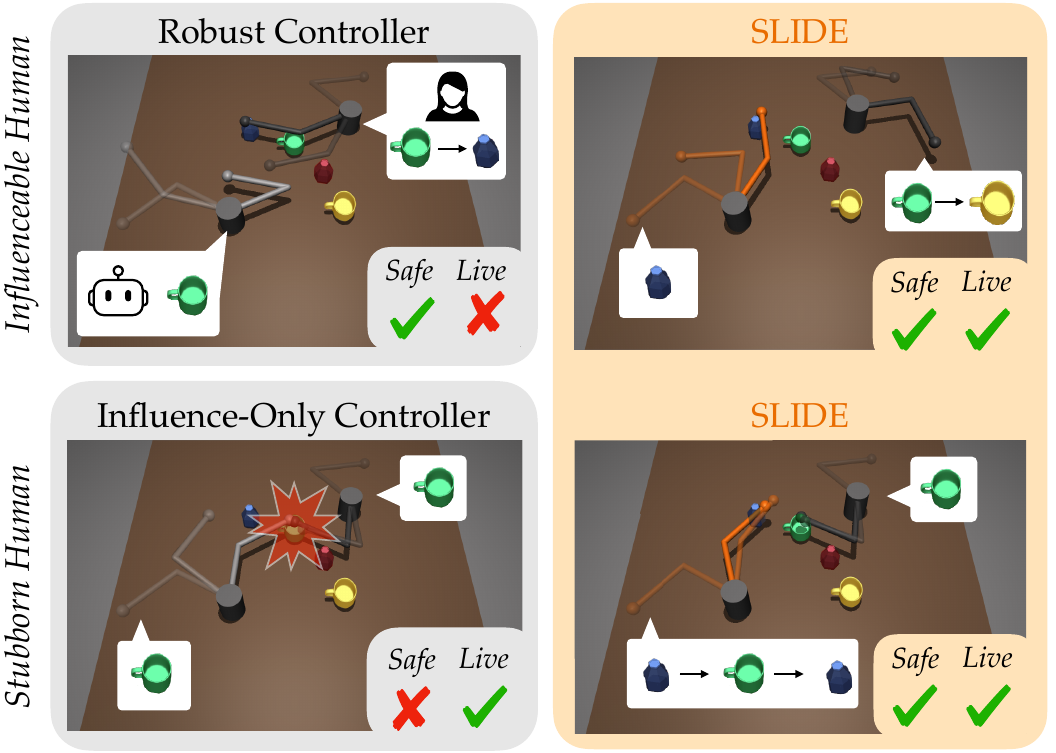}
    \caption{Both human and robot arms want to reach their desired objects on the table, but they don't know who is going for which object. \textbf{Top Row}: The human's desired object can be influenced by the robot. Using a influence-\textit{unaware} safety shield the robot can stay safe, but fails to reach its own object (not \textit{live}). 
    With our method (SLIDE) the robot influences the human's goal and safely reaches its object. \textbf{Bottom Row:} The human never changes their desired object. Naive influence-aware robot controllers are over-confident  and collide.
    SLIDE recognizes that this can be unsafe and chooses a different goal for the robot, staying safe and live. }
    \label{fig:front_fig}
    \vspace{-1.5em}
\end{figure}

Whether intentional or not, influence underlies many multi-agent interactions, from nudging into someone's lane while driving to merge faster, to grabbing your favorite bottle first so that your partner has to get a different one (\figref{fig:front_fig}, top right). 
While exploiting such influence can enable agents like robots to be more efficient, it can also lead to unsafe outcomes: if you quickly reach for your favorite mug but your partner doesn't adapt fast enough or is unwilling to change, then you can cause a collision (\figref{fig:front_fig}, bottom left). 

In this work, we seek to enable robots to \textit{safely} influence during human-robot interactions.
However, we face two challenges, one from the human modeling perspective and the other from the robot control perspective.   
On one hand, it is difficult to hand-design a model that captures the complexity of how people can be influenced by the robot's behavior. 
On the other hand, the robot actions that are maximally influential are also often those than can lead to states of irrecoverable failure where \textit{no} safe robot action exists.

To tackle this complexity, we pose a novel robust reach-avoid dynamic game between the human and robot.  
First, we take inspiration from data-driven trajectory forecasting \cite{tolstaya2021identifying} and inform the human's behavior in the dynamic game via a deep conditional behavior prediction (CBP) model. 
With CBPs, the robot can learn implicit patterns in the responses of the human \textit{conditioned} on other agents' future behavior. 
Second, we solve the reach-avoid game in the joint physical and robot belief space. 
This enables the robot to reach its goal while staying robust to uncertainty over the human's future behavior, instead of always trusting what the conditional model predicts for a short horizon.
Finally, to solve this high-dimensional game offline, we adopt approximate reach-avoid reinforcement learning solvers \cite{hsu2023isaacs} that have recently shown promise in scaling to high-dimensional systems \cite{hu2023deception, nguyen2024gameplay}. 

With our framework, called \textbf{SLIDE} (Safely Leveraging Influence in Dynamic Environments), we can compute robot policies that exploit influence to maximize efficiency (i.e., liveness) while staying robust to uncertainty and minimizing safety violations (right, \figref{fig:front_fig}).
Through extensive simulations in a 39-dimensional human-robot collaborative manipulation scenario,  we show that SLIDE is less conservative than prior safe control approaches while staying safe even in the presence of out-of-distribution human behavior.

\section{Related Work}

\para{Application: Safe Collaborative Manipulation}
We ground our approach in human-robot collaborative manipulation tasks, which are common in industrial manufacturing \cite{mukherjee2022survey} and will become common in home environments (e.g. cooking and cleaning tasks) as robotic assistants grow more popular \cite{abou2020systematic}. Ensuring \textit{safety} is critical in these domains \cite{villani2018survey}, has been studied by prior works \cite{liu2016algorithmic, singletary2021safety, landi2019safety, liu2022safe}, notably via energy-function-based safe control (e.g. Control Barrier Functions \cite{ames2019control} and the Safe Set Algorithm \cite{liu2014control}). 
We are motivated by this domain, but focus specifically on the safety challenges stemming from \textit{influence} in collaborative manipulation.

\para{Modeling Human-Robot Influence} 
While many works have studied robot influence on humans via expressions or appearance \cite{saunderson2019robots, siegel2009persuasive, rae2013influence, admoni2017social}, we focus on physical action. 
One line of prior work has modeled influence \textit{implicitly} by learning the dynamics of a latent representation of the collaborator's strategy \cite{xie2021learning, wang2022influencing, parekh2022rili}. 
Notably, these methods do not consider influence within a single interaction, only between interactions. Other work has considered modeling influence \textit{explicitly} by predicting the future actions of other agents in driving scenarios using learned models \cite{tolstaya2021identifying, ngiam2021scene, huang2023conditional} or stackelberg games \cite{sadigh2016planning, sagheb2023towards, fisac2019hierarchical, schwarting2019social} and in collaborative manipulation scenarios \cite{bestick2017implicitly, kedia2024interact, pandya2024towards}. 
In contrast, we focus specifically on the \textit{safety} problems that emerge when robots take influential actions. 


\para{Embedding Human Models in Safe Robot Control} 
While foundational works treat the human as a disturbance \cite{fisac2015reach}, recent works embed predictive human models into safe controllers to reduce their conservativeness. 
This has been done by adapting a dynamics and uncertainty model of the human online \cite{liu2015safe, liu2022safe, pandya2022safe, pandya2024multimodal}, or via limiting the forward reachable sets of a human based on predictions \cite{fisac2018probabilistically, bajcsy2019scalable, bajcsy2020robust, nakamura2023online}. We instead consider a \textit{backward} reachability approach, similar to \cite{tian2022safety, hu2023deception} so that the robot can explicitly reason about what it can do to prevent unsafe outcomes, thereby further reducing conservativeness. 
However, we are focused on collaborative manipulation and leverage data-driven conditional behavior prediction to capture robot influence on people. 






\section{Background: Reach-Avoid Dynamic Games}
\label{sec:background}


Our method is rooted in reach-avoid dynamic games \cite{fisac2015reach}. 
While there are several ways to solve these games, we leverage Hamilton-Jacobi (HJ) reachability analysis \cite{mitchell2005time} which is a safe control technique compatible with general nonlinear systems, control constraints and disturbances, and is associated with a suite of numerical synthesis techniques \cite{mitchell2005toolbox, fisac2019bridging, bansal2021deepreach}. Here we provide a brief overview of HJ reachability (see \cite{bansal2017hamilton} for a review).

\para{Human \& Robot Dynamics} 
We model the human and robot as the two players in the dynamic game. 
Let the robot state be denoted by $x_\robot^t \in \stateSpaceR$ and the human as $x_\human^t \in \stateSpaceH$. The human-robot system state is $x^t=(x_\robot^t, x_\human^t) \in \stateSpace$ and evolves via the deterministic discrete-time dynamics $x^{t+1} = f(x^t, \aR^t, \aH^t)$
where the robot and human control inputs are denoted by $\aR^t \in \ctrlSetR$ and $\aH^t \in \ctrlSetH$ respectively. 

\para{Reach-Avoid Games via HJ Reachability} 
HJ reachability computes a backward reachable tube (BRT), $\stateSpace^\shield \subset \stateSpace$, which characterizes the set of initial states from which the robot is guaranteed to reach a desired target set while also avoiding a set of failure states, despite the best effort of an adversary. 
It also synthesizes a corresponding reach-avoid robot policy, $\fallback$.
Let the target set $\mathcal{T} := \{x \mid l(x) \leq 0\}$ and failure set $\mathcal{F}:=\{x \mid g(x) > 0\}$ be encoded via the Lipschitz-continuous margin functions $l(\cdot)$ and $g(\cdot)$. 
For robustness, the reach-avoid game models the robot as attempting to stay safe while reaching the goal, and the human as a virtual adversary who attempts to thwart this. 
Solving the game amounts to computing the value function characterized by the fixed-point Isaacs equation \cite{isaacs1954differential}:
\begin{equation}
    \label{eq:hji-vi}
    \begin{split}
        V(x) = \max \Big\{ g(x), \min \Big\{ l(x), 
        \min_{\aR\in\ctrlSetR} \max_{\aH\in\ctrlSetH} V(x^+) \Big\} \Big\},
    \end{split}
\end{equation}
where $x^+ = f(x, \aR, \aH)$ is the next state. 
The sub-zero level set of the value function $\safeSet = \{x \mid V(x) \leq 0\}$ encodes our desired set of states (i.e., BRT) from which there exists a robot control signal that can reach the target set without ever entering failure, despite a worst-case adversary. 
The corresponding optimal robot policy $\fallback$ can be obtained via:
\begin{equation}
    \fallback(x) = \argmin_{\aR\in\ctrlSetR} \max_{\aH\in\ctrlSetH} V(x^+).
    \label{eq:safe-policy}
\end{equation}

\section{SLIDE: Safely Leveraging Influence in Dynamic Environments}
\label{sec:method}

\begin{figure*}
    \centering
    \includegraphics[width=1\textwidth]{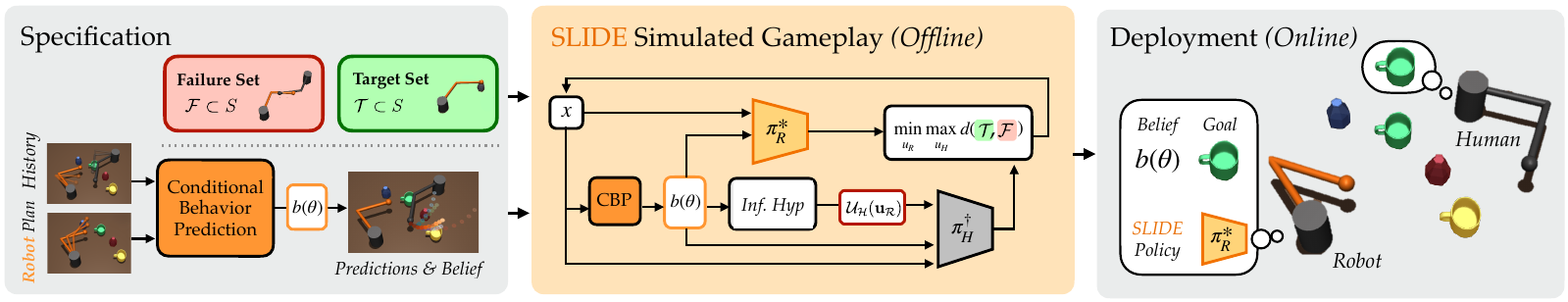}
    \caption{\textbf{SLIDE Framework.} (left) Before solving the reach-avoid game, we specify the target set (goal locations), failure set (collisions), and a conditional behavior prediction (CBP) model that can predict the human's future trajectory conditioned on the robot's future plan. (center) During simulated gameplay, the SLIDE policy, $\fallback(x_e)$, is trained against a simulated human adversary, $\policy_\human^\dagger(x_e)$ whose control bounds are informed by the CBP model. (right) Online, the robot uses its robust SLIDE policy to safely influence against \textit{any} human. }
    \label{fig:system_diagram}
    \vspace{-0.2in}
\end{figure*}




The reach-avoid formulation from Section~\ref{sec:background} establishes the backbone of our approach. 
However, applying it directly to influencing human-robot interactions would face two challenges. 
First, the model of the human is far too pessimistic by treating them as a best-effort adversary (thus making $\fallback$ overly conservative). 
Second, it assumes that the robot can never learn about the human during interaction and thus can never decrease (or increase) its uncertainty about their future behavior. 
To extend this mathematical framework to safely account for human-robot influence, we introduce two key modifications which we detail below: (1) a conditional behavior prediction model of human influence, and (2) a belief-space formulation of reach-avoid games. 

\para{Modeling Influence via Conditional Behavior Prediction} 
Let $\aHtraj:=[\aH^t,\ldots,\aH^{t+k}]^\top$ be a trajectory of human actions for a horizon of length $k$ and $\aRtraj:=[\aR^t,\ldots,\aR^{t+k}]^\top$ represent the same for the robot. 
Our approach leverages a pre-trained conditional behavior prediction (CBP) model which 
outputs a multimodal distribution over human action trajectories $\aHtraj$ conditioned on the robot's future plan $\aRtraj$: $\mathcal{P} := P(\aHtraj \mid x^t, \aRtraj; \theta)$ where $\theta\in\Theta$ represents the mode of the distribution. Our approach is agnostic to the particular form of the CBP model, only requiring that the model outputs $M$ discrete modes (i.e. $|\Theta|=M$) and their associated probabilities $p_\theta$. Note that learning a Gaussian Mixture Model (which adheres to this form) is already common among existing trajectory forecasting models \cite{tolstaya2021identifying, shi2022motion, salzmann2020trajectron++}. The belief of the robot then is the $M$ modes and their associated probabilities: $b_\robot^t := \{(\theta_i,p_{\theta_i})\}_{i=1}^M$.

Our core idea is that the CBP model enables us to inform the actions we expect the human to take within the reach-avoid game (i.e., $\ctrlSetH$ in Eqn.~\ref{eq:hji-vi}) while also capturing the influence the robot has over this action set. 
Specifically, for each predicted behavior mode $\theta \in \Theta$, we construct the set of $\delta-$likely trajectories under the predicted distribution $\mathbb{\ctrlSett}_\human(\aRtraj; \theta) := \{\aHtraj \mid P(\aHtraj \mid x^t, \aRtraj; \theta) \geq \delta\}$. We construct our \textit{influence-informed control bounds} for the human by taking the time-wise minimum and maximum actions: 
\begin{equation}
    \ctrlSetH(\aRtraj; \theta) := [\min_{t\in\{t,\ldots t+k\}} \mathbb{\ctrlSett}_\human, \max_{t\in\{t,\ldots t+k\}} \mathbb{\ctrlSett}_\human].
\end{equation}
Similar to \cite{hu2023deception}, we have an \textit{inference hypothesis}: at each time $t$, the robot's belief $b_\robot^t$ will assign at least probability $\epsilon$ to the human's next action $\aH^t$. Mathematically, we assume the human's action will belong to the inferred control bound during reach-avoid analysis:
\begin{equation}
    \ctrlSetH(\aRtraj) := \bigcup_{\theta_i \in \{\theta \mid b_\robot^t(\theta) \geq \epsilon\}} \ctrlSetH(\aRtraj; \theta_i).
    \label{eq:inference-hypothesis}
\end{equation}
Here, $\epsilon\geq 0$ is a hyperparameter that captures the reliability of the trajectory forecasting model.
Now that we have a influence-aware human model we can inject into our reach-avoid game, we have to contend with the challenge that the robot's belief---and therefore Eqn.~\ref{eq:inference-hypothesis}---changes over time. 

\para{Belief-Space Reach-Avoid Games}
To enable the robot to account for uncertainty in the human's future behavior, we modify the reach-avoid problem from Eqn.~\eqref{eq:hji-vi} by \textit{extending} the state to include both the physical state ($\state$) and the robot's \textit{belief state} ($b_\robot(\theta)$). 
Here, the robot's belief state is precisely the Gaussian mixture model over possible behavior modes $\theta$ output by the CBP.  
Note that at each timestep the robot re-generates its predictions of the human behavior given new observations. 
This induces the belief-space dynamics: 
\begin{equation}
    b_\robot^{t+1} = f_L(b_\robot^t, x^t, \aR, \aH).
\end{equation}
Note that in this work, $f_L$ is implicitly modeled by subsequent calls to the forecasting model during reach-avoid reinforcement learning.

Finally, let the extended state be $x^t_e=(x^t, b_\robot^t)$ and the extended physical-belief dynamics to be $x_e^{t+1} = F(x_e^t, \aR, \aH)$. 
Putting our model from Eqn.~\ref{eq:inference-hypothesis}, we obtain a modified fixed-point Isaacs equation:
\begin{equation}
    \begin{split}
        V({\color{dark_orange} x_e}) = \max \Big\{ g({\color{dark_orange} x_e}), \min \Big\{ l({\color{dark_orange} x_e}), \min_{\aR} \max_{\aH  {\color{dark_orange} \in\ctrlSetH(\aRtraj)} } V( {\color{dark_orange} x_e^+}) \Big\} \Big\}.
    \end{split}
    \label{eqn:slice-value}
\end{equation}
where ${\color{dark_orange} x_e^+} = {\color{dark_orange} F}({\color{dark_orange} x_e}, \aR, \aH)$ is the next joint physical and belief state.
The differences from the standard Isaacs equation in Eqn.~\ref{eq:hji-vi} is highlighted in {\color{dark_orange}orange}: this is the inclusion of the extended state and the adversary's control bound being a function of the ego's nominal long-term plan.
 

\para{\textit{Offline:} Reach-Avoid Reinforcement Learning Solution}
Our problem in Eqn.~\eqref{eqn:slice-value} quickly becomes computationally intractable with traditional level-set methods \cite{mitchell2005toolbox, bui2022optimizeddp} due to the dimensionality of the extended state. 
However, we build on the recent ISAACS reach-avoid reinforcement-learning based solver \cite{hsu2023isaacs} to solve a time-discounted version of the safety value function \eqref{eqn:slice-value}. 
We use a soft actor-critic formulation to train the value function critic $V(\cdot)$ and two actor networks, $\policy_\robot^*(\cdot)$ and $\policy_\human^\dagger(\cdot)$, representing the optimal robot and simulated human adversary control policies. 
To enforce our inference hypothesis in Eqn.~\eqref{eq:inference-hypothesis}, we project the actions of the adversary agent's policy to be within the inferred control bound $\ctrlSetH(\aRtraj)$. 
At each step during training, the robot's future plan $\aRtraj$ is computed by querying a nominal policy to drive the robot towards the closest goal. 


\para{\textit{Online}: Safe and Influencing Control}
Online, we deploy the trained reach-avoid policy $\policy_\robot^*(x_e)$ directly via solving Eqn.~\eqref{eq:safe-policy}. We input the current state $x$ and query the trajectory prediction model for the belief state $b_\robot$ at each timestep. 


\section{Experimental Setup}
\label{sec:experiments-setup}

\para{Task: Tabletop Object Reaching} We deploy our framework in a scenario where a human and robot arm need to reach their desired objects on the table, but they do not know who is going for what object. 
They must choose objects without colliding into each other. 
We instantiate four objects on the table, two mugs and two bottles (shown in \figref{fig:front_fig}) denoted by $g_i \in \mathcal{G}$ in Cartesian space (i.e. end-effector goals) and via discrete semantic class, $c: \mathcal{G} \rightarrow \mathbb{W}$.

\para{Human-Robot System Dynamics} We model the human and robot as 2-link planar manipulators (i.e. no gravity) and each agent $i\in\{\human,\robot\}$'s dynamics are:
\begin{equation}
    M_i(q_i)\Ddot{q}_i + C_i(q_i,\dot{q}_i)\dot{q}_i = B_iu_i,
\end{equation}
where the state $q_i\in \qSpace_i$ consists of the joint angles, $M_i$ is the inertia matrix, $C_i$ captures Coriolis forces, and $B_i$ represents how the control input affects the system. 
The agents' actions $\ctrl_i\in\ctrlSet_i$ are the joint torques bounded by a box constraint: $\ctrlSet_i:=\{\ctrl \mid \ctrl_i^{min} \leq \ctrl \leq \ctrl_i^{max}\}$. 
We model the robot as having larger control authority than the human.
The configuration space states $q_i$ are transformed by an observation map $\mathcal{O}_i :\qSpace_i \rightarrow \mathbb{R}^{n_i}$ where $n_i$ is the dimensionality of the observation for agent $i$. 
The physical state input to the model is the concatenation of observations of both agents: $\stateSpace := \mathcal{O}_\robot(\qSpace_\robot) \times \mathcal{O}_\human(\qSpace_\human)$. 

\para{Multi-Arm Interaction Data Generation} 
To create our conditional behavior prediction model, we need a dataset of multi-agent interactions to learn from. 
Unlike in autonomous driving, where many large, diverse, multi-agent datasets already exist \cite{ettinger2021large, caesar2020nuscenes, chang2019argoverse}, these types of large datasets are underrepresented in collaborative manipulation. 
We thus choose to create our own synthetic dataset of multi-agent interactions so that we can control how the other agent is influenced can cleanly analyze the effect of our influence-aware safe control policy.
Our simulated human always tries to choose an object with a different type than the robot. We simulate this by having the human keep their own belief over the goal that robot is currently choosing, $b_\human^t(g) \propto P(u_\robot^t \mid q_\robot^t, g) b_\human^{t-1}(g)$, which gets updated online via Bayes' Rule. 
The human changes their goal to be the least likely robot goal based on $b_\human^t$ if the robot's most likely goal both has probability $> 0.3$ and has the same semantic class as the human's goal.
The dataset is generated by randomly sampling initial goals for the two agents and rolling out for fixed horizon of 15 seconds.  

\begin{figure*}
    \centering
    \includegraphics[width=0.95\textwidth]{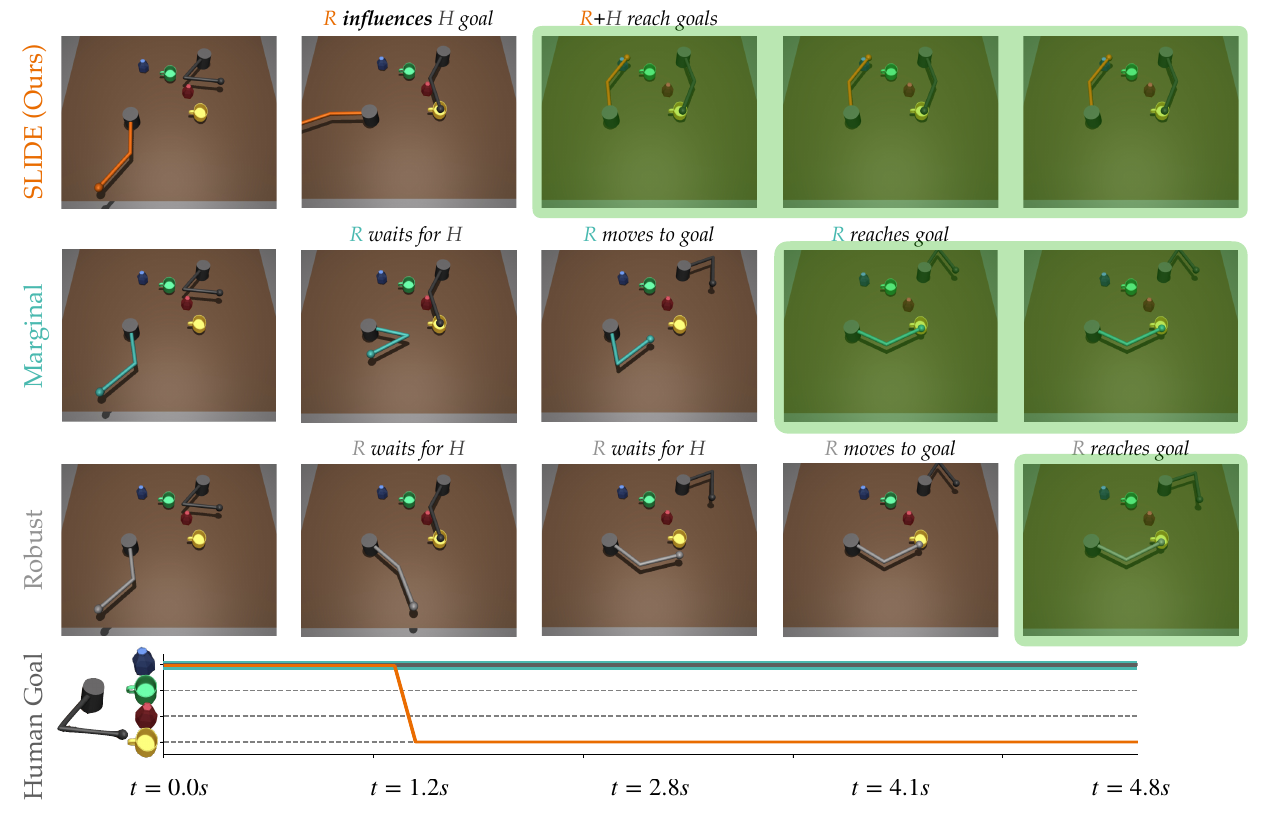}
    \caption{\textbf{Closed-Loop Simulations.} \ours, \marginal and \robust policies starting from the same initial condition. SLIDE confidently understands that the human will be influenced to move out of its way as it chooses the blue bottle and reaches the fastest (the human changes its mind from the blue bottle to the yellow mug at $t=1.2s$). Marginal-RA waits until the human is out of its way and chooses the yellow mug. Robust-RA stays cautious even as the human is moving towards a different goal and finishes last. 
    }
    \label{fig:traj-rollout}
\end{figure*}



\para{Methods} We compare two human prediction models: marginal and conditional (CBP). We also compare our method, \ours, with four other robot policies: \nosafety (computed torque control), the Safe Set Algorithm (\ssa) \cite{liu2014control}, Robust reach-avoid (\robust) \cite{hsu2023isaacs}, and Marginal reach-avoid (\marginal) \cite{hu2023deception}. We use a naive version of SSA to understand the effect of adding simple safe control. The robust policy solves the same reachability problem but without including a prediction model, and the marginal policy has the same structure as SLIDE but its prediction model is not conditioned on the robot's future plan. 


\para{Metrics} For trajectory prediction models, we measure the average displacement error (ADE), final displacement error (FDE), and size of the inferred control bound $|\ctrlSetH|$. For closed-loop simulations of the agents, we measure: 1) collision rate (safety) 2) task completion rate (liveness) and 3) completion time (average trajectory length). 

\para{Training: Prediction Models for Manipulation} The prediction models are both 3-layer MLPs\footnote{We found this simple architecture to be sufficient for our task, but our method is agnostic to the complexity of the model.} with hidden sizes of 256 neurons and output the parameters of a Gaussian Mixture Model (GMM) to predict the human's actions for the next 1 second. Both models' input consists of a 1-second history of both agents' end effector positions, the goal positions $\Theta$, and the goals' semantic classes $c(\theta_i)$ as a flattened vector. The CBP model additionally takes in a 2-second future plan of the robot's end effector position. The models are trained with the sum of two loss function terms (similar to \cite{tolstaya2021identifying}): 1) negative log-likelihood and 2) MSE of most-likely GMM mode. 


\para{Training: Safety Value Function} The actor and safety critic networks are 4-layer MLPs with a hidden size of 256 and the actor outputs a Guassian distribution for the action of the robot. The input to the networks consists of the 24D observed state (plus 15D extended belief state for marginal and SLIDE methods). The belief state consists of the mode means and mixture weights of the trajectory predictor's GMM output.
The policies are trained using one NVIDIA GeForce RTX 4090 GPU and the RL environment is run on 16 parallel threads on an AMD Ryzen Threadripper 7960X CPU. The wall-clock time for training the robust policy is approximately 5 hours, the marginal policy is approximately 28 hours, and SLIDE is 61 hours. This includes time for pretraining the ego and adversary agents, following the training scheme laid out by \cite{hsu2023isaacs}. 


\section{Experimental Results}
\label{sec:experiments-results}

\subsection{Influence-Aware vs. Influence-Unaware Safety}

We first study the closed-loop performance of \ours's controller, which safely exploits influence, compared to alternative safe controllers described in \sref{sec:experiments-setup}. 


\begin{figure}
    \centering
    \includegraphics[width=\columnwidth]{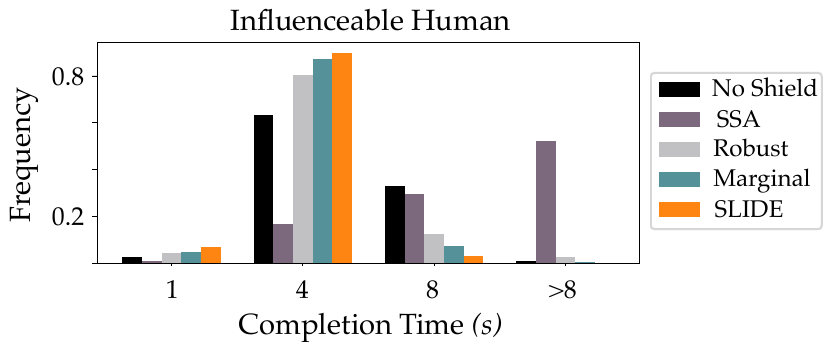}
    \caption{\textbf{Closed-loop Completion Times.} Histogram of completion times for all methods interacting with the influenceable human model. 
    \ours has the highest frequency of short trajectories, while \ssa and \robust have the highest incidence of timing out.}
    \label{fig:time_hist}
    \vspace{-0.5em}
\end{figure}

\begin{table}[ht]
    \centering
    \resizebox{\columnwidth}{!}{%
    \begin{tabular}{l|c c c c}
        \toprule
         & \multicolumn{1}{c}{Collision rate} & \multicolumn{1}{c}{Completion rate} & \multicolumn{1}{c}{Completion Time (s)} \\
        \midrule
        \nosafety & $28.5\%$ & $71.5\%$ & $3.5\pm 1.8$ \\
        \ssa & $19.1\%$ & $52.3\%$ & $8.9 \pm 4.7$ \\
        \robust & $1.4\%$ & $97.0\%$ & $2.6 \pm 2.1$ \\
        \marginal & $1.5\%$ & $98.0\%$ & $2.5\pm 1.3$ & \\
        \ours (ours) & $1.9\%$ & $98.1\%$ & $1.9\pm 0.8$ \\
        \bottomrule
    \end{tabular}
    }
    \caption{\textbf{Closed-loop Results. } Failure rate, completion rate and completion time for all methods over 1,000 randomized trials. The task is incomplete if the robot does not reach any goal within 15 seconds. \ours is able to reach its goal most often and in the shortest time without becoming significantly less safe.}
    \label{tab:policy_eval}
    \vspace{-0.1in}
\end{table}

\para{Results: Quantitative}
\tabref{tab:policy_eval} shows all methods' performance. As expected, \robust has the smallest collision rate and all safe controllers have a lower collision rate than \nosafety.
However, what \robust and \marginal gain in safety, they lose in completion rate and time. In contrast, \ours has the highest completion (liveness) rate while having a comparable collision rate to the robust and marginal baselines. 
Moreover, \ours allows the robot to reach goals significantly faster: e.g., $24\%$ faster than \marginal. 
We further plot the histogram of completion times  in \figref{fig:time_hist}. 
We see that \ours almost never times out, while other methods do. 
We note that \ssa collides and times out the most out of the safe controllers. 
The reason for this is because the safety index for \ssa was hand-tuned (as proposed in \cite{liu2014control}), so it does not always respect the system's control limits. If we used some additional safety index synthesis techniques \cite{zhao2023safety} to properly account for the control bounds, the performance would likely approach that of \robust, though without the goal-reaching policy built-in.

\para{Results: Qualitative} We visualize \ours and \robust's closed-loop trajectories in \figref{fig:traj-rollout}. 
\robust initially keeps the robot arm far away from all goals, waiting for the human to move out of the way before reaching for a mug.
In contrast, \ours immediately recognizes that it can influence the human's target object (bottom, \figref{fig:traj-rollout}). It switches to reach the bottle, ensuring the human picks a cup, stays out of its way, and enables the robot to complete the task faster. 

\subsection{Ablation: When Does Modeling Influence Matter?}
Next, we study when it matters that we use influence-aware human models for safe control. 
We ablate if the robot uses a \textit{conditional} or \textit{marginal} prediction model (i.e. no conditioning on robot's future plan).
We compare both the performance of the predictors and their effect on the learned reach-avoid policies. 


\begin{table}[ht]
    \centering
    \resizebox{\columnwidth}{!}{%
    \begin{tabular}{l|c|| c c c}
        \toprule
         & \multicolumn{1}{c||}{All Data (14,000)} & \multicolumn{1}{c}{Interactive Data (2,457)} & \multicolumn{1}{c}{Non-Interactive Data (11,543)} \\
        \midrule
        \textbf{Marginal} & $0.002$ ($0.12$) & $0.007$ ($0.40$) & $0.001$ ($0.07$) \\
        \textbf{CBP} & $0.001$ ($\textbf{0.09}$) & $0.007$ ($\textbf{0.30}$) & $0.0007$ ($\textbf{0.04}$) \\
        \bottomrule
    \end{tabular}
    }
    \caption{\textbf{Open-Loop Prediction Error.} Average (ADE) and Final Displacement Error (FDE) of marginal and CBP predictors.}
    \label{tab:cbp-v-marg}
\end{table}

\begin{table}[ht]
    \centering
    \resizebox{\columnwidth}{!}{%
    \begin{tabular}{l|c|| c c c}
        \toprule
         & \multicolumn{1}{c||}{All Data (14,000)} & \multicolumn{1}{c}{Interactive Data (2,457)} & \multicolumn{1}{c}{Non-Interactive Data (11,543)} \\
        \midrule
        \textbf{Marginal} & [12.3, 12.7] & [13.8, 13.6] & [11.9, 12.5] \\
        \textbf{CBP} & [7.8, 8.3] & [12.7, 12.9] & [6.7, 7.3] \\
        \bottomrule
    \end{tabular}
    }
    \caption{\textbf{Inferred $\ctrlSet_\human$ Size.} Average inferred control bound size from \marginal ($|\ctrlSetH|$) and \ours ($|\ctrlSetH(\aRtraj)|$). Entries shown per ctrl. dimension and the max. dyn. feasible range is 20.}
    \vspace{-1.5em}
    \label{tab:cbp-v-marg-bounds}
\end{table}

\para{Approach} In these experiments, the human agent acts according to the data distribution in \sref{sec:experiments-setup}. 
We evaluate the predictors on a dataset  $\mathcal{D}$ of 100 held-out trajectories, each 15 seconds long, resulting in 14,000 data points\footnote{We discount predictions in the last 1 second of each trajectory since this is length of the prediction horizon.}.

\begin{figure}[t!]
    \centering
    \includegraphics[width=1\columnwidth]{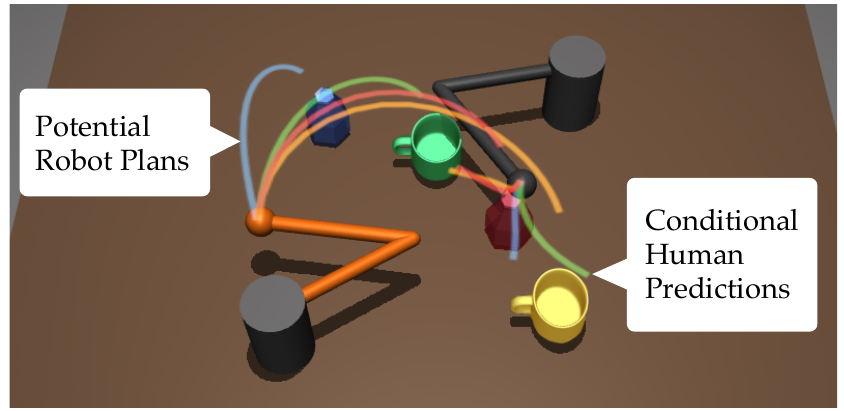}
    \caption{\textbf{Conditional Behavior Predictions.} Most-likely mode of \ours CBP model given different future robot plans. Each robot plan has a corresponding human prediction in the same color. The prediction is highly dependent on the robot's plan and captures the idea that the human will change goals to a different semantic class.}
    \label{fig:cbp_predictions}
    \vspace{-1em}
\end{figure}

\para{Open-Loop Results: Quantitative \& Qualitative} 
\tabref{tab:cbp-v-marg} shows ADE and FDE results. Averaged across the held-out dataset, ADE looks very similar for both marginal and CBP, with CBP performing better on FDE (left, \tabref{tab:cbp-v-marg}). 
However, only a \textit{subset} of all trajectories in this held-out dataset exhibit highly interactive scenarios. 
Thus, we further decompose dataset $\mathcal{D} = \mathcal{D}_I \cup \mathcal{D}_{\neg I}$ into data points where influence is happening $\mathcal{D}_I$ (2,457 data points) and those where it is not $\mathcal{D}_{\neg I}$ (11,543 data points). 
This is done by tracking the timesteps where the human's goal changes: if it does, then this and directly adjacent timesteps are added to $\mathcal{D}_{I}$. 
For data in $\mathcal{D}_{I}$, the CBP model has a significantly lower FDE compared to the marginal trajectory predictor (center column, \tabref{tab:cbp-v-marg}), indicating that the CBP matters to predict long-term behavior in highly interactive scenarios.
\figref{fig:cbp_predictions} shows the CBP model that \ours uses as a function of the robot's plan. 
Each robot plan results in distinct human predictions. This capability (not present in a marginal prediction model) gives the \ours policy the freedom to learn which goals to move towards to best influence the human to move out of its way.
We further study the implications of each model on the inferred control bounds, $\ctrlSetH$, that we use during reach-avoid computations (\sref{sec:method}). 
This will directly tell us how the prediction model will affect the policy training loop, since this determines the set of allowable actions for the adversary agent. 
In \tabref{tab:cbp-v-marg-bounds} we show the  size of the inferred control bounds. The CBP action bounds enlarge at interactive states, but do not expand as much as the marginal predictor's bounds.
On the full dataset, the CBP model results in a smaller control bound on average. 
This implies that \ours's downstream policy (which uses the CBP) will be able to exploit the human's influence and thus choose less conservative actions. 

\para{Closed-Loop Results: Quantitative \& Qualitative} Finally, we study the closed-loop safety and liveness performance of of the \ours policy and the \marginal policy. 
In \tabref{tab:policy_eval}, we see that both methods have a similar task completion rate (\ours is 98.1\% while \marginal is 98.0\%). 
However, the main difference comes in their completion \textit{time}: \ours completes the task in $1.9s$ compared to \marginal at $2.5s$. This implies that our \ours policy exploits the influence it has on the human to maximize task completion efficiency. 
Qualitatively, we see that \ours chooses the top (blue) bottle (top row, \figref{fig:traj-rollout}) to influence the human into picking to the bottom (yellow) mug, making the two agents stay out of each others' way. 
In contrast, the \marginal policy chooses the goal closest to itself (middle row, \figref{fig:traj-rollout}), but then must wait for the human to move out of the way before reaching the yellow mug.

\subsection{How Robust is SLIDE to Out-of-Distribution Humans?}

\begin{figure}
    \centering
    \includegraphics[width=\columnwidth]{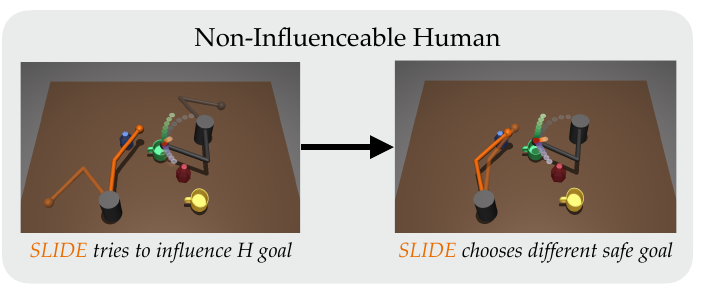}
    \caption{Shows an interaction between the \ours controller and a stubborn (non-influenceable) human model. The modes of the GMM output by the CBP prediction model are visualized as dots from the human's end effector. 
    }
    \label{fig:no_influence}
\end{figure}

Finally, we study how robust the \ours policy is 
when interacting with human behavior that is out-of-distribution (OOD) from the conditional behavior prediction model. We measure the same closed-loop metrics described in \sref{sec:experiments-setup}. 

\para{Approach}
We ablate the deployment-time human model (right, \figref{fig:system_diagram}) to be OOD relative to the prediction model training data.  
Specifically, we change the human to be \textbf{non-influenceable} (also referred to as \textit{stubborn}) and \textbf{adversarial} (where the human always chooses the goal it thinks the robot is likely moving towards\footnote{Note that this is different from the optimal disturbance policy $\pi_\human^\dagger$ trained in simulated gameplay.}). 

\begin{table}[t!]
    \centering
    \resizebox{\columnwidth}{!}{%
    \begin{tabular}{l|c c c c}
        \toprule
         & \multicolumn{1}{c}{Collision rate} & \multicolumn{1}{c}{Completion rate} & \multicolumn{1}{c}{Completion Time (s)} \\
        \midrule
        \textbf{Influenceable Human} & $1.9\%$ & $98.1\%$ & $1.9\pm 0.8$ \\
        \textbf{Stubborn Human (OOD)} & $1.8\%$ & $98.1\%$ & $2.0\pm 0.9$ \\
        \textbf{Adversarial Human (OOD)} & $3.8\%$ & $96.2\%$ & $1.9\pm 0.7$ \\
        \bottomrule
    \end{tabular}
    }
    \caption{\textbf{Closed-Loop OOD Human Results.} \ours interacts with in-distribution and out-of-distribution (OOD) humans over 1,000 randomized trials. 
    Stubborn human never changes their goal. Adversarial human always chooses the goal it thinks the robot is moving towards. SLIDE is relatively robust to the stubborn human, but starts to degrade against the adversarial human.}
    \label{tab:policy_eval_ood}
    \vspace{-1em}
\end{table}

\para{Results: Quantitative \& Qualitative} 
Results are shown in \tabref{tab:policy_eval_ood}
for 1,000 random initial conditions. 
We see that when interacting with a stubborn human, \ours generally performs similarly to the in-distribution human model. 
We hypothesize this is precisely because of the \textit{inference hypothesis} from \sref{sec:method}: instead of blindly trusting the output, we allow the simulated opponent agent to take any $\delta-$likely action from $\epsilon-$likely modes. 
This means as long as the modes of the prediction model have some coverage of the agent's true behavior, the learned policy will still be robust to it. 
We can see an example of this in \figref{fig:no_influence}. 
At first, \ours tries to influence the human to switch to the red bottle goal. 
After a few timesteps, however, \ours backs off and chooses a different (safe) goal (the blue bottle) to complete the task.
When interacting with the \textbf{adversarial human} that always chooses the goal it thinks the robot is moving towards, we can see that \ours starts to collide more often (bottom row, \tabref{tab:policy_eval_ood}). 
This makes sense since the human sometimes takes actions that are in direct opposition to the predictor's training data, which our inference hypothesis is unable to account for. 
However, it is a promising sign that there is not a catastrophic degradation in the collision rate.
Future work should investigate, for example, the use of a \textit{hybrid} approach which switches between the \robust controller and \ours depending on the prediction quality observed during deployment. 
\section{Conclusion}
\label{sec:conclusion}
In this work, we enable robots to safely influence humans. 
We pose and solve a new reach-avoid dynamic game (called SLIDE) that (1) accounts for influence via the use of a data-driven conditional behavior prediction model and (2) accounts for the robot's ability to learn online via a belief-state. 
In simulations, we find that SLIDE can accomplish tasks significantly faster than prior safe control approaches, and remains relatively robust in the face of out-of-distribution human interactions. 
Future work should investigate calibrating the prediction model \cite{lindemann2023safe} as well as performing post-verification on the approximate reach-avoid value function \cite{lin2024verification, yang2024scalable} for stronger safety assurances. 



\section*{Acknowledgment}
This work is supported by NSF Award \#2246447. RP is also funded by the NSF GRFP. 

\bibliographystyle{IEEEtran}
{\footnotesize
\bibliography{references}
}

\end{document}